\documentclass{article}
\usepackage{spconf,amsmath,amssymb,graphicx,bbm,mathtools,multirow,hhline,url}
\DeclareMathOperator*{\argmax}{arg\,max}

\title{Modular Hybrid Autoregressive Transducer}
\name{\begin{tabular}{c}
Zhong Meng, Tongzhou Chen, Rohit Prabhavalkar, Yu Zhang, Gary Wang, Kartik Audhkhasi, \\ Jesse Emond, Trevor Strohman, Bhuvana Ramabhadran, W. Ronny Huang, \\ Ehsan Variani, Yinghui Huang, Pedro J. Moreno \end{tabular}
}
\address{Google LLC., USA}
\copyrightnotice{978-1-6654-7189-3/22/\$31.00~\copyright2023 IEEE}
\begin{document}
\ninept
\maketitle

\begin{abstract}
%Text-only adaptation of a transducer model remains a challenging task since it has no clearly separated acoustic model (AM) and language model (LM) and estimates correlated blank and label distributions.
%A hybrid autoregressive transducer (HAT) models label and blank posteriors using two separate distributions, and is a promising candidate for text adaptation.
%In this work, we propose a modular HAT (MHAT) that has structurally separated label and blank decoders to predict label and blank distributions, respectively, with a shared acoustic encoder. The outputs of encoder and label decoder are directly projected to AM and internal LM scores, respectively, before being added together to estimate the label posteriors. 
%We train MHAT with an internal LM loss and an HAT loss to ensure that its internal LM becomes a standalone neural LM and the rest model components are able to work with it to predict accurate label and blank posteriors. Therefore, the MHAT internal LM can be effectively adapted to text-only data using any neural LM adaptation methods 
%without contaminating the blank distribution or the AM scores. More importantly, ILMA of MHAT leads to a more effective LM fusion than internal LM subtraction-based methods.
%Experimented with multi-domain production data, we show that an MHAT adapted by 100B sentences achieves up to 12.4\% and 21.5\% relative word error rate reductions from 400K-hour trained HAT, respectively, with and without additional LM fusion.

Text-only adaptation of a transducer model remains challenging for end-to-end speech recognition since the transducer has no clearly separated acoustic model (AM), language model (LM) or blank model. In this work, we propose a modular hybrid autoregressive transducer (MHAT) that has structurally separated label and blank decoders to predict label and blank distributions, respectively, along with a shared acoustic encoder. The encoder and label decoder outputs are directly projected to AM and internal LM scores and then added to compute label posteriors. We train MHAT with an internal LM loss and a HAT loss to ensure that its internal LM becomes a standalone neural LM that can be effectively adapted to text. Moreover, text adaptation of MHAT fosters a much better LM fusion than internal LM subtraction-based methods. On Google's large-scale production data, a multi-domain MHAT adapted with 100B sentences achieves relative WER reductions of up to 12.4\% without LM fusion and 21.5\% with LM fusion from 400K-hour trained HAT.

\end{abstract}
\begin{keywords}
Speech recognition, text-only adaptation, hybrid autoregressive transducer
\end{keywords}
\section{Introduction}
\label{sec:intro}

End-to-end (E2E) automatic speech recognition (ASR) has achieved state-of-the-art performance by directly mapping speech to word sequences through a single neural network. The most popular E2E models include connectionist temporal classification \cite{graves2006connectionist, hannun2014deep}, recurrent neural network transducer (RNN-T) \cite{graves2012sequence, jain2019rnn, sainath2020streaming,  li2020developing}, and attention-based encoder-decoder (AED) models \cite{chorowski2015attention, chan2016listen, chiu2018state}.
However, E2E models can easily overfit to the source-domain training data, resulting in performance degradation in a mismatched target domain.
Numerous methods have been explored to adapt ASR models, such as regularization methods \cite{kld_yu, meng2019asa, l2_liao, meng2020lvector, lhuc}, teacher-student learning \cite{li2014learning, meng2018adversarial, manohar2018teacher, meng2019conditional}, transformation methods \cite{lhn, tan2015cluster, sc_abdel}, and adversarial learning \cite{grl_shinohara, meng2018speaker, grl_serdyuk, dsn_meng}. Nevertheless, all these adaptation methods require audio data when applied to E2E models \cite{ochiai2018speaker, meng2019speaker, meng2019domain}.
To overcome this, one promising approach is to adapt the E2E model using text-only data such that we can take advantage of orders of magnitude more unpaired text than the audio-speech paired utterances.

Language model (LM) fusion is a simple yet effective approach to achieve text-only adaptation. 
%For LM fusion, an external LM is trained separately using target-domain text. 
In \cite{hannun2014deep,gulcehre2015on}, a shallow fusion was proposed to linearly combine the E2E model score with the LM score in the log domain at each step of beam search inference. Density ratio method \cite{mcdermott2019density,kanda2016maximum,variani2015gaussian} 
improves shallow fusion by subtracting a source-domain LM score from the shallow fusion score. More recently, a hybrid autoregressive model (HAT) \cite{variani2020hybrid} was proposed to
facilitate an estimation of the internal LM score (ILME), which is then subtracted from the shallow fusion score during inference, significantly improving ASR in the target domain.
% estimate an internal LM score and subtract it from the shallow fusion score at each step of beam search.
Similar ILME-based fusion was proposed for other E2E models in \cite{meng2021ilme, zeyer2021librispeech}. However, LM fusion methods require running an additional LM during inference, leading to increased run-time model parameters, higher computational cost, and longer decoding time.

An alternative solution is to synthesize speech from adaptation text using a text-to-speech (TTS) system and fine-tune the E2E model using synthesized audio-transcript pairs \cite{peyser2019improving,huang2020rapid,chen2020improving}. Although these techniques improve ASR performance, training TTS models and generating high-quality speech are both computationally expensive. Moreover, additional clean supervised training data is required to train a reliable multi-speaker TTS model.  During adaptation, a transducer loss is optimized for RNN-T and HAT with frame-level input and forward-backward computation, requiring much higher computational cost than text data fine-tuning with a cross-entropy loss. 
% Without the need for re-train the E2E model, LM fusion allows for a fast and flexible text-only adaptation.

To overcome the weakness of both LM fusion and TTS-based approaches, directly fine-tuning the E2E model with text-only data was proposed and has shown to be effective for domain adaptation. \cite{pylkkonen2021fast} learns an additional LM output component 
% using source-domain data 
to regularize the text-only fine-tuning of an RNN-T predictor. However, training this extra LM output layer increases computational cost and adaptation time. To overcome this, an internal LM adaptation (ILMA) method \cite{meng2021ilma} was proposed by fine-tuning the joint network of RNN-T with text after internal LM training \cite{variani2020hybrid,meng2021ilmt}. ILMA was shown to outperform shallow fusion with much fewer run-time parameters. However, ILMA was designed for RNN-T models where no clear separation of acoustic model (AM), internal LM and blank model exists. Adapting the internal LM of RNN-T may distort the way other model components affect the RNN-T output, counteracting the benefits from text-only adaptation. To address this, \cite{chen2022factorized} factorizes blank and label predictions by introducing an additional blank predictor. Trained with RNN-T and LM losses, the factorized transducer (FT) learns a standalone LM internally that can be fine-tuned with adaptation text for label prediction. 
% However, the factorized transducer was trained to minimize an RNN-T loss where blank and vocabulary logits are normalized together via a single \texttt{Softmax} which mismatches the LM adaptation loss where only the vocabulary logits are normalized. 
However, in FT, blank and label logits are normalized together via a single \texttt{Softmax} during training and testing while the label logits are normalized only over themselves during text-only adaptation. HAT \cite{variani2020hybrid} and \cite{zeyer2020new} model label and blank posteriors with two separate distributions and are natural candidates to compensate for the adaptation and testing mismatch.

In this work, we combine the advantages of HAT and FT, and propose a \emph{modular HAT (MHAT)}. Inheriting from HAT, MHAT computes the blank posterior individually using a \texttt{Sigmoid} and applies a separate \texttt{Softmax} normalization over only label logits, eliminating the mismatch between adaptation and testing. MHAT is trained with HAT loss with an additional internal LM loss. Inspired by FT, MHAT has separate label and blank decoders. The acoustic encoder works with the label and blank decoders to predict the label and blank posteriors, respectively. The encoder and blank decoder are directly projected to generate AM and internal LM scores of labels which are then combined at the output layer. All of these ensure that the internal LM scores are computed independently from AM scores or the blank posterior; thus, the internal LM can be effectively adapted to text-only data without affecting other MHAT components. During ILMA, our loss differs from the one used in FT in that it includes Kullback-Leibler divergence (KLD) \cite{kullback1951information} regularization of internal LM to prevent the degradation of source-domain word error rate (WER). Furthermore, we show that ILMA can foster a much better LM fusion with MHAT. Experimented on both public and Google's production datasets, MHAT with ILMA can achieve up to 25.6\% and 14.4\% relative WER reductions with and without additional LM fusion, respectively.

\section{Related Work}
\subsection{Hybrid Autoregressive Transducer}
\label{sec:hat}
An E2E model estimates the posterior distribution $P(\mathbf{Y} |
\mathbf{X};\theta_\text{E2E})$ over sequences of output labels $\mathbf{Y}=\{y_1, \ldots,
y_U\}$ given a sequence of input speech features $\mathbf{X}=\{\mathbf{x}_1,
\ldots, \mathbf{x}_T\}$, where $y_u \in \mathcal{V}, u = 1, \ldots, U$, and $\mathbf{x}_t
\in \mathbbm{R}^{d_x}, t = 1, \ldots, T$. $\mathcal{V}$ is the set of all possible labels, e.g., word pieces, etc. $y_0$ is the start of sentence token. 

HAT \cite{variani2020hybrid} is a time-synchronized model that predicts a conditional distribution $P(\mathbf{\tilde{Y}} |
\mathbf{X};\theta_\text{HAT})$ over blank-augmented token sequences, i.e., alignments, $\mathbf{\tilde{Y}} = \{\tilde{y}_1, \ldots, \tilde{y}_{T+U}\}$, where $\tilde{y}_i \in \mathcal{V}\: \cup\:\texttt{<b>}, i = 1, \ldots, T + U$, and $\texttt{<b>}$ is a blank that determines an alignment between the input speech and output token sequences. The label sequence posterior $P(\mathbf{Y} |
\mathbf{X};\theta_\text{HAT})$ is then computed by marginalizing over all possible alignments that can collapse into $\mathbf{Y}$.
\begin{align}
P(\mathbf{Y} | \mathbf{X};\theta_\text{HAT}) = \sum_{\mathbf{\tilde{Y}} \in \mathcal{B}^{-1}(\mathbf{Y})} P(\mathbf{\tilde{Y}} | \mathbf{X};\theta_\text{HAT})
\end{align}
where $\mathcal{B}$ is a function that removes blanks from the alignment $\mathbf{\tilde{Y}}$.
% $\mathbf{\tilde{Y}}$ is aligned with the token and feature sequences $\mathbf{Y}$ and $\mathbf{X}$ as $\left(\tilde{y}_i, y_{u_i}, \mathbf{x}_{t_i}\right)^{U + T}_{i=1}$,
%\begin{align}
%    \left(\tilde{y}_i, y_{u_i}, x_{t_i}\right)^{U + T}_{i=1}
%\end{align}
% where the index $i$ in $\mathbf{\tilde{Y}}$ is mapped to the index $u_i$ in $\mathbf{Y}$, and the index $t_i$ in $\mathbf{X}$.

HAT consists of an acoustic encoder, a decoder and a joint network. The encoder transforms the input speech features $\mathbf{X}$ into acoustic embedding vectors $\mathbf{F} = \{\mathbf{f}_1, \ldots, \mathbf{f}_T\}, \; \mathbf{f}_t \in \mathbbm{R}^{d_f}$. The decoder takes in previous labels to generate the current label embedding $\mathbf{g}_u \in \mathbbm{R}^{d_g}$ as follows
\begin{align}
\mathbf{F} & = \text{Encoder}(\mathbf{X}), \label{eqn:encoder} \\
\mathbf{g}_u & = \text{Decoder}(\mathbf{Y}_{0:u - 1}). \label{eqn:decoder}
\end{align}
% Both the acoustic and label embeddings are vectors with the same dimension $d$.

The joint network combines the acoustic and label embedding vectors via a feed-forward network, and then estimates label and blank posteriors separately with different distributions.
% by applying \texttt{Sigmoid} and \texttt{Softmax}, respectively. 
The blank posterior given the alignment history is computed based on a conditional Bernoulli distribution as
\begin{align}
b_{t, u} & = P(\tilde{y}_{t + u} = \texttt{<b>}|\mathbf{X}_{1:t}, \mathbf{\tilde{Y}}_{0:t + u - 1}) \nonumber \\
& = \text{Sigmoid}[\mathbf{w}^\intercal \phi(\mathbf{W}_1 \mathbf{f}_t + \mathbf{W}_2 \mathbf{g}_u)] \label{eqn:blank_posterior}
\end{align}
where $\mathbf{W}_1 \in \mathbbm{R}^{d_h\times d_f}$ and $\mathbf{W}_2 \in \mathbbm{R}^{d_h\times d_g}$ are projection matrices.
% project acoustic and label embeddings, respectively, into $d_h$-dimensional hidden vectors. 
$\phi(\cdot)$ is a non-linear function, e.g., Tanh, ReLU. The vector $\mathbf{w} \in \mathbbm{R}^{d_h}$ followed by \texttt{Sigmoid} transforms the non-linear output into a scalar between 0 and 1. Note that in this paper, we omit the bias vectors in all linear transforms for simplicity.

The label posteriors given previous speech features and labels are computed as
\begin{align}
P(y_u|\mathbf{X}_{1:t}, \mathbf{Y}_{0:u - 1}) = \text{Softmax}[\mathbf{W} \phi(\mathbf{W}_1 \mathbf{f}_t + \mathbf{W}_2 \mathbf{g}_u)] \label{eqn:label_posterior}
\end{align}
where the same non-linear output is transformed to a $|\mathcal{V}|$-dimensional logit vector through the matrix $\mathbf{W} \in \mathbbm{R}^{|\mathcal{V}|\times d_h}$ before Softmax normalization.
The label posteriors given the alignment history is therefore the above label posteriors weighted by a non-blank probability \begin{align}
 P(\tilde{y}_{t + u} &= y_u|\mathbf{X}_{1:t}, \mathbf{\tilde{Y}}_{0:t + u - 1}) = (1 - b_{t,u}) P(y_u|\mathbf{X}_{1:t}, \mathbf{Y}_{0:u - 1}). \label{eqn:label_alignment_posterior}
% & \hspace{-3pt} = [1 - P(\tilde{y}_{t + u} \hspace{-2pt} = \hspace{-1pt}\texttt{<b>}|\mathbf{X}_{1:t}, \mathbf{\tilde{Y}}_{0:t + u - 1})] P(y_u|\mathbf{X}_{1:t}, \mathbf{Y}_{0:u - 1}). \label{eqn:label_alignment_posterior}
\end{align}
%With blank and label posteriors given alignment history. The alignment posterior can be computed by substituting Eqs. \eqref{eqn:blank_posterior} and \eqref{eqn:label_alignment_posterior} into Eq. \eqref{eqn:} below
%\begin{align}
%P(\mathbf{\tilde{Y}} | \mathbf{X};\theta_\text{HAT}) = \sum_{i=1}^{T+U-1} P(\tilde{y}_i | \mathbf{X}, \mathbf{\tilde{Y}}_{0:i - 1};\theta_\text{HAT})
%\end{align}
% Since text-only adaptation can only affect label posteriors, the separate estimation of blank and label posteriors makes HAT a promising candidate for internal LM adaptation without affecting blank distribution.
Since our goal is to adapt the internal LM probability using text without affecting the blank distribution, HAT with separate blank and label predictions is a promising candidate for ILMA.
% Since the text data contains only non-blank tokens, 

HAT is trained to minimize the summed negative log posteriors of all label sequences on the training corpus $\mathcal{D}_\text{T}$: 
\begin{align}
    \hspace{-2pt}\mathcal{L}_{\text{HAT}}(\theta_\text{ILM}; \mathcal{D}_\text{T}) = -\sum_{(\mathbf{X}, \mathbf{Y}) \in \mathcal{D}_\text{T}} \log P(\mathbf{Y}|\mathbf{X};\theta_\text{HAT}). \label{eqn:hat_loss}
\end{align}

\subsection{External LM Fusion with HAT}
% via the E2E training with paired audio and transcripts. The internal LM
The E2E model implicitly learns an internal LM $P(\mathbf{Y}; \theta_\text{E2E})$ that characterizes the label sequence distribution given $\mathbf{\theta}_\text{E2E}$.
%The internal LM probability of a token sequence $\mathbf{Y}$ is
%\begin{align}
%    & P(\mathbf{Y};\theta_\text{E2E}) = \prod^{U}_{u=1}P(y_u|\mathbf{Y}_{0:u-1};\theta_\text{E2E}), \label{eqn:e2e_ilm}
%%    & = \sum_{X} \prod^{U}_{u=1} P(y_u|\mathbf{X}, \mathbf{Y}_{0:u-1};\theta^\text{S}_\text{E2E}) P(\mathbf{X} | \mathbf{Y}_{0:u-1};\theta^\text{S}_\text{E2E}). \label{eqn:e2e_cond_ilm}
%\end{align}
The internal LM probability is estimated as the E2E model output after zeroing out the acoustic embeddings $\mathbf{F}$ \cite{variani2020hybrid, meng2021ilme}. For HAT, the internal LM consists of the decoder followed by the joint network and is computed as follows
% For a transducer, the internal LM probability is the model output after feeding the previous tokens $\mathbf{Y}_{0:u-1}$ through the internal LM, i.e., the prediction network followed by the joint network
\begin{align}
P(y_u|\mathbf{Y}_{0:u - 1};\theta_\text{ILM}) = \text{Softmax}[\mathbf{W} \phi(\mathbf{W}_2 \mathbf{g}_u)],
\end{align}
where $\theta_\text{ILM}$ denotes the internal LM parameters.

% The ASR performance of HAT can be improved by integrating an external LM during inference. 
When an external LM is available for inference, we subtract the internal LM log probability from the log-linear combination between the E2E model probability and the external LM probability $P(\mathbf{Y}; \hat{\theta}_\text{LM})$ at each step of beam search to alleviate the effect of source-domain internal LM and facilitate a more effective fusion with the target-domain external LM \cite{meng2021ilme}. The optimal label sequence $\hat{\mathbf{Y}}$ is obtained as follows
% The optimal token sequence $\hat{\mathbf{Y}}$ is obtained by maximizing the following criterion
\begin{align}
    \hat{\mathbf{Y}} = \argmax_{\mathbf{Y}} & \left[\log P(\mathbf{Y}|\mathbf{X}; \hat{\theta}_\text{HAT}) + \lambda_\text{E} \log P(\mathbf{Y}; \hat{\theta}_\text{LM}) \right. \nonumber \\
                     & \left. \quad - \lambda_\text{I} \log P(\mathbf{Y}; \hat{\theta}_\text{ILM}) \right], \label{eqn:ilme}
\end{align}
where $\lambda_\text{E}$ and $\lambda_\text{I}$ are the weights for external and internal LM probabilities, respectively. $\hat{\theta}_\text{HAT}$ and $\hat{\theta}_\text{LM}$ denote well-trained HAT and external LM parameters, respectively.

\section{Modular HAT}
To adapt the internal LM of HAT using text-only data, we fine-tune the decoder and joint network of HAT to minimize a cross-entropy loss.
Although HAT predicts the blank and label posteriors separately, both predictions are made given the same decoder output $\mathbf{g}_u$ as in Eqs. \eqref{eqn:blank_posterior} and \eqref{eqn:label_posterior}. Updating the shared decoder parameters in Eq. \eqref{eqn:decoder} will contaminate the blank distribution because text-only data does \emph{not} contain any blank tokens. To overcome this, we structurally decouple the blank prediction from the label prediction by replacing the single decoder with a separate label decoder and a separate blank decoder as inspired by \cite{chen2022factorized}. The label decoder output is combined with the encoder output to predict the label posteriors while the blank decoder works with the \emph{shared} encoder to predict the blank posterior. With this design, the blank distribution will not be affected when we adapt only the label decoder to the text-only data.

In a HAT model, the speech features make their impact on the final output through an encoder followed by a joint network. However, the joint network is shared by both the encoder and the label decoder to compute the HAT output and is trained with both speech and text data. Adapting the joint network with text-only data will distort the way an encoder output contributes to the HAT output. To circumvent this, 
% we remove the joint network from the label decoder output. Instead, 
we directly project the acoustic and label embeddings to $|\mathcal{V}|$-dimensional acoustic model (AM) and LM logits, respectively, and then apply a \texttt{LogSoftmax} function to compute the AM and internal LM log probabilities, respectively. Finally, we add the AM and internal LM scores together and apply a \texttt{Softmax} normalization to compute the label probability. 
% or essentially test speech 

The above two modifications modularize the HAT model: First, the encoder and label decoder compute AM and internal LM probabilities independently without affecting one another, and the two probabilities are not combined until the output layer. 
% predicting label posteriors. 
Secondly, the blank posterior is estimated independently of the internal LM probability by a separate blank decoder and the shared encoder. Therefore, we name this model a \emph{modular HAT (MHAT)}. MHAT ensures that its internal LM is a standalone neural LM that is easily adaptable to text-only data without distorting the blank distribution or the encoder's contribution to the final output. 

\subsection{MHAT Architecture}

\begin{figure}[htpb!]
	\centering
	\includegraphics[width=0.9\columnwidth]{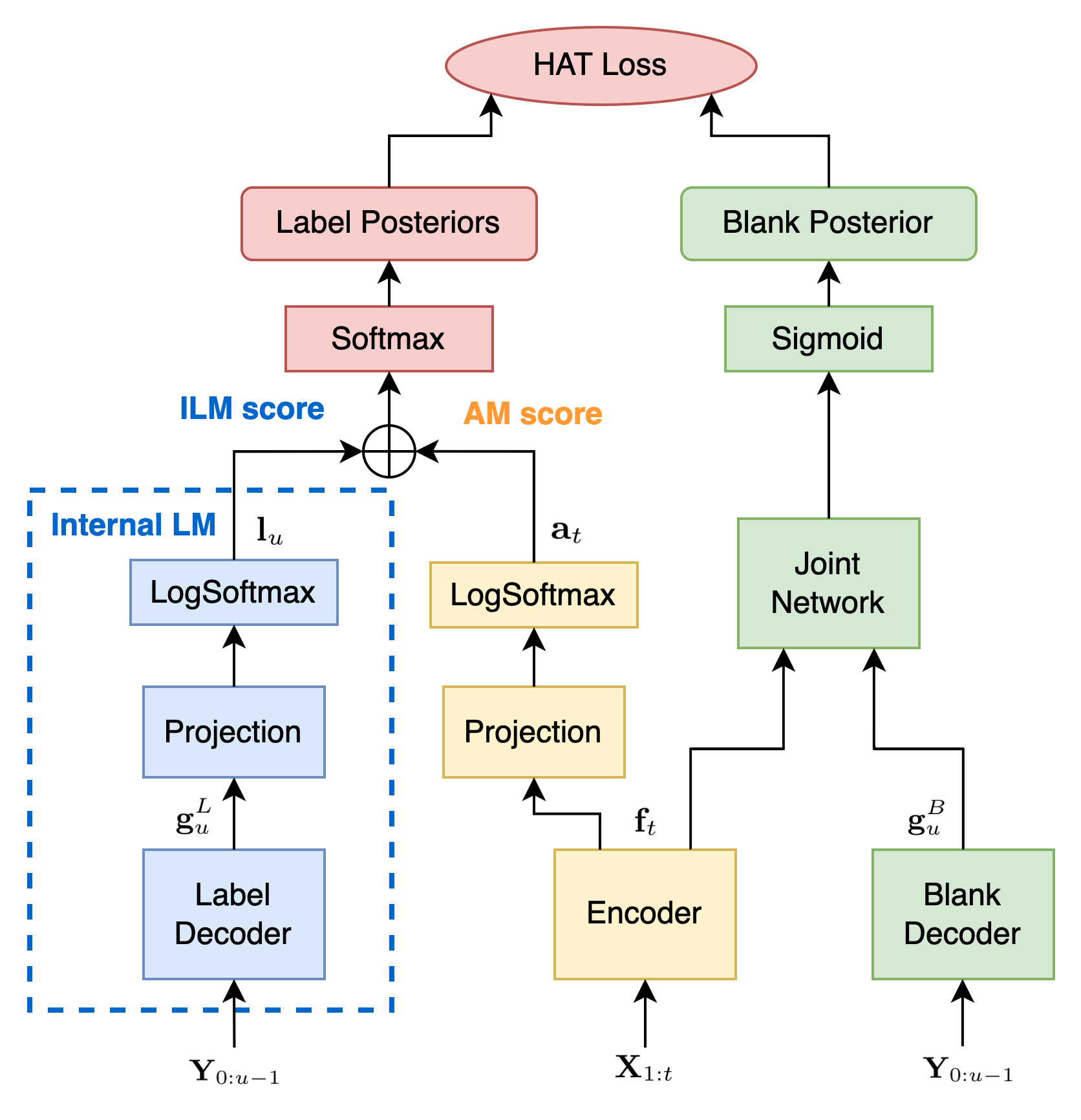}
    \vspace{-10pt}
    \caption{The architecture of a modular HAT (MHAT). The encoder and blank decoder are directly projected to AM and internal LM scores which are then combined at the output layer to compute label posteriors. The blank posterior is predicted by the encoder and the blank decoder followed by a joint network. The label and blank distributions are estimated with \texttt{Softmax} and \texttt{Sigmoid}, respectively. Only the internal LM in the blue dashed box will be adapted to text-only data.}
	\label{fig:mhat}
\end{figure}

As shown in Fig. \ref{fig:mhat}, MHAT consists of a shared acoustic encoder, a blank decoder and a label decoder. Identical to the HAT encoder, the shared MHAT encoder maps a sequence of speech features $\mathbf{X}$ into a sequence of acoustic embeddings $\mathbf{F}$ as in Eq. \eqref{eqn:encoder}. Both label and blank decoders take the same previous labels as the input to generate their respective current label embeddings below 
\begin{align}
\mathbf{g}^\text{B}_u = \text{BlankDecoder}(\mathbf{Y}_{0:u - 1}), \\
\mathbf{g}^\text{L}_u = \text{LabelDecoder}(\mathbf{Y}_{0:u - 1}),
\end{align}
where $\mathbf{g}^\text{B}_u \in \mathbbm{R}^{d^\text{B}_g}$ and $\mathbf{g}^\text{L}_u \in \mathbbm{R}^{d^\text{L}_g}$ are label embeddings of the blank and label decoders, respectively. 
%$d^B_g$ and $d^L_g$ are the dimensions of a blank decoder embedding and a label encoder embedding, respectively.
The blank posterior given the alignment history is obtained in a similar way as for a standard HAT:
\begin{align}
P(\tilde{y}_{t + u} & = \texttt{<b>}|\mathbf{X}_{1:t}, \mathbf{\tilde{Y}}_{0:t + u - 1}) \nonumber \\
& = \text{Sigmoid}[\mathbf{w}^\intercal \phi(\mathbf{W}_1 \mathbf{f}_t + \mathbf{W}_2 \mathbf{g}^\text{B}_u)]. \label{eqn:blank_posterior_2}
\end{align}
The acoustic embedding from encoder is projected and then normalized to be a $|\mathcal{V}|$-dimensional (dim) vector of AM log probabilities $\mathbf{a}_t$. The label embedding from label decoder is projected and normalized to be a $|\mathcal{V}|$-dim vector of internal LM log probabilities $\mathbf{l}_u$.
\begin{align}
\mathbf{a}_t & = \text{LogSoftmax}(\mathbf{W}_3 \mathbf{f}_t) \\
\mathbf{l}_u & = \text{LogSoftmax}(\mathbf{W}_4 \mathbf{g}^\text{L}_u) \label{eqn:ilm_mhat}
\end{align}
where $\mathbf{W}_3 \in \mathbbm{R}^{|\mathcal{V}|\times d_f}$ and $\mathbf{W}_4 \in \mathbbm{R}^{|\mathcal{V}|\times d^\text{L}_g}$ are projection matrices.
% and $\mathbf{b}_3 \in \mathbbm{R}^{|\mathcal{V}|}$ and $\mathbf{b}_4 \in \mathbbm{R}^{|\mathcal{V}|}$ are projection biases.
The AM and internal LM log probabilities are added together and then normalized by a \texttt{Softmax} to compute the label posteriors given previous speech features and labels as follows
\begin{align}
P(y_u|\mathbf{X}_{1:t}, \mathbf{Y}_{0:u - 1}) = \text{Softmax}\left(\mathbf{a}_t + \mathbf{l}_u \right) \label{eqn:label_posterior_2}
\end{align}
The label posteriors given the alignment history is then computed by substituting Eq. \eqref{eqn:label_posterior_2} into Eq. \eqref{eqn:label_alignment_posterior}. 

Although important, the blank decoder normally needs much fewer parameters than the label decoder since it performs a much simpler task of predicting only one blank token. 
% estimates the posterior of only one token which is a much simpler task than predicting $|\mathcal{V}|$ label posteriors performed by the label decoder. 
We show in the experiments that MHAT can perform as well as HAT with only 1\% of the model parameters assigned to the blank decoder.

Intuitively, the label decoder followed by $\mathbf{W}_4$ forms the internal LM of MHAT since it is the only component that contributes to the internal LM probability. Unlike other E2E models where the internal LM probability can only be \emph{approximated} by zeroing out the encoder output (see Proposition 1 in \cite[Appendix A]{variani2020hybrid}), the \emph{accurate} internal LM of MHAT can be computed by Eq. \eqref{eqn:ilm_mhat} which can potentially improve the ILME-based LM fusion in Eq. \eqref{eqn:ilme}.
% For a transducer, the internal LM probability is the model output after feeding the previous tokens $\mathbf{Y}_{0:u-1}$ through the internal LM, i.e., the prediction network followed by the joint network

\subsection{MHAT Training}
\label{sec:mhat_train}
In a modular HAT, the AM and internal LM log probabilities are computed independently through an encoder plus $\mathbf{W}_3$ and a label decoder plus $\mathbf{W}_4$, respectively. The AM and internal LM scores are added and then re-normalized by a \texttt{Softmax} to compute the label posteriors at the output layer. Furthermore, the blank posterior is completely independent of the internal LM probability because it is computed with the shared encoder and a separate blank decoder. Therefore, adapting the label decoder and $\mathbf{W}_4$, i.e., the internal LM, using text-only data will not distort the AM probabilities or the blank distribution.

To ensure that the internal LM of MHAT is adaptable to text, we need to make it a standalone neural LM with low perplexity on text-only data and, just as importantly, to make the other parts of MHAT capable of collaborating with such a standalone neural LM to predict label and blank posteriors. To achieve these, we train MHAT to minimize an internal LM loss in addition to the HAT loss using audio-transcript pairs from scratch as follows
\begin{align}
    & \hspace{-3pt} \mathcal{L}_{\text{MHAT}}(\theta_\text{MHAT}; \mathcal{D}_\text{T}) = \mathcal{L}_{\text{HAT}}(\theta_\text{MHAT}; \mathcal{D}_\text{T}) + 
    \alpha \mathcal{L}_{\text{ILM}}(\theta_\text{ILM}; \mathcal{D}_\text{T}), \label{eqn:ilmt}
\end{align}
where the internal LM loss is the summed negative log internal LM probabilities over the \emph{transcripts} of the training corpus $\mathcal{D}_\text{T}$ as follows
\begin{align}
    \hspace{-2pt}\mathcal{L}_{\text{ILM}}(\theta_\text{ILM}; \mathcal{D}_\text{T}) &= -\sum_{\mathbf{Y} \in \mathcal{D}_\text{T}} \sum^{U}_{u=1}\log P(y_u|\mathbf{Y}_{0:u-1};\theta_\text{ILM}) \nonumber \\
    &= -\sum_{\mathbf{Y} \in \mathcal{D}_\text{T}} \sum^{U}_{u=1}\log l_{u,y_u}, \label{eqn:ilm_loss_train}
\end{align}
and $\alpha > 0$ is the weight of the internal LM loss. $\theta_\text{ILM}$ equals to label decoder parameters plus $\mathbf{W}_4$ for MHAT. $l_{u,y_u}$ is the $y_u$ th dimension of $\mathbf{l}_u$
% and $P(y_u|\mathbf{Y}_{0:u-1};\theta_\text{ILM})$ is obtained by Eq. \eqref{eqn:ilm_mhat}. 
% This procedure is similar to the internal LM training \cite{variani2020hybrid, meng2021ilmt} of other transducer models.

With a proper $\alpha$, MHAT can be trained to perform similarly as being trained with an HAT loss alone, but have a much lower-perplexity internal LM. 
% the perplexity of its internal LM is significantly reduced. 
This implies that the internal LM has become a standalone neural LM being able to collaborate with the encoder and blank decoder to achieve comparable ASR performance. 

\subsection{Internal LM Adaptation of MHAT}
% To adapt MHAT using text-only data, any neural LM adaptation methods can be applied to adapt the internal LM of MHAT. 
As a standalone neural LM, the internal LM of MHAT can be adapted to text-only data with any neural LM adaptation methods.
The most intuitive solution is to fine-tune the internal LM, i.e, the label decoder plus $\mathbf{W}_4$, to minimize a simple cross-entropy internal LM loss $\mathcal{L}_{\text{ILM}}(\theta_\text{ILM};\mathcal{D}_\text{A})$,
%\begin{align}
%    & \hspace{-2pt}\mathcal{L}_{\text{ILM}}(\theta_\text{ILM};\mathcal{D}_\text{A}) = - \hspace{-2pt} \sum_{\mathbf{Y} \in \mathcal{D}_\text{A}} \sum^{U}_{u=1}\log P(y_u|\mathbf{Y}_{0:u-1};\theta_\text{ILM}). \label{eqn:ilm_loss_adapt}
%\end{align}
where $\mathcal{D}_\text{A}$ is the set of text-only adaptation data.

In many occasions, we want to improve the ASR performance on target-domain data while maintaining the performance on source-domain data. To achieve this, we regularize the internal LM output during ILMA by minimizing an additional KLD between the adapted and unadapted internal LMs.
The ILMA loss therefore becomes the following 
\begin{align}
    & \mathcal{L}_{\text{ILMA}}(\theta_\text{ILM}; \mathcal{D}_\text{A}) = (1 - \rho) \mathcal{L}_{\text{ILM}}(\theta_\text{ILM}; \mathcal{D}_\text{A}) \nonumber \\
    & - \rho \hspace{-2pt} \sum_{\mathbf{Y} \in \mathcal{D}_\text{A}} \sum_{u = 1}^U \sum_{v \in\mathcal{V}} P( v|\mathbf{Y}_{0:u-1};\theta^*_\text{ILM}) \log P( v|\mathbf{Y}_{0:u-1};\theta_\text{ILM}), \nonumber \\
\end{align}
where $\rho \in [0, 1]$ is the regularization weight and $\theta^*_\text{ILM}$ is the internal LM parameters before adaptation.

% MHAT well separates the prediction of internal LM probability from that of the AM probability and the blank posterior through structurally separated blank decoder and label decoder, and direct projections from the encoder output and the label decoder output to label posteriors. 
As explained in Section \ref{sec:mhat_train}, with ILMA, only the internal LM of MHAT is adapted to target-domain text data while the AM probability and the blank distribution remain unchanged, effectively improving the ASR performance in the target domain. By taking token-level text as the input and minimizing a cross-entropy loss, ILMA facilitates a much faster adaptation of the MHAT model than TTS-based approaches which take frame-level speech inputs and minimize a transducer loss. Most importantly, without any external LMs during inference, ILMA enables a much faster decoding with lower computational cost than LM fusion methods.

\subsection{External LM Fusion with MHAT}
To further improve the ASR accuracy on target-domain data, we integrate an external LM into the \emph{internal-LM-adapted (ILMA-ed)} MHAT during inference.
We sum the log probabilities generated by MHAT and the external LM at each step of beam search and obtain the optimal decoded label sequence $\hat{\mathbf{Y}}$ as follows
% The optimal token sequence $\hat{\mathbf{Y}}$ is obtained by maximizing the following criterion
\begin{align}
    \hat{\mathbf{Y}} = \argmax_{\mathbf{Y}} & \left[\log P(\mathbf{Y}|\mathbf{X}; \hat{\theta}_\text{MHAT}) + \lambda_\text{E} \log P(\mathbf{Y}; \hat{\theta}_\text{LM}) \right] \label{eqn:mhat_lm_fusion}
\end{align}
where $\hat{\theta}_\text{MHAT}$ is the model parameters of the ILMA-ed MHAT.

Note that, different from HAT \cite{variani2020hybrid} or the ILME-based fusion \cite{meng2021ilme}, we do \emph{not} subtract the internal LM score from the combined MHAT and LM score because the internal LM of MHAT has been adapted to target-domain text which allows for a better fusion with the target-domain external LM. We show in our experiments that the external LM fusion with an ILMA-ed MHAT leads to significant WER reduction over an ILMA-ed MHAT even if the external LM is trained with the same text used for adaptation. It also consistently outperforms the external LM fusion with an internal-LM-\emph{subtracted} HAT or MHAT in terms of lower WER. 

\section{Experiments}
We verify the effectiveness of the proposed MHAT on two experimental setups: cross-domain adaptation on public datasets and multi-domain adaptation on Google's production datasets. In the cross-domain setup, we adapt an MHAT trained with Librispeech data \cite{panayotov2015librispeech} to the training transcripts of WSJ, TED-LIUM \cite{rousseau2012ted} and Common Voice \cite{ardila2019common} datasets, respectively. In the multi-domain setup, we train an MHAT with cascaded encoders using multi-domain transcribed production data and adapt it to large-scale multi-domain text-only data.

\subsection{Cross-Domain Adaptation}
\subsubsection{Dataset}
\label{sec:public_data}
% We evaluate a source-domain E2E model on target-domain test data by integrating an external LM trained with the text-only data in the target domain. We use the RNN-T or AED trained with 30K-hour multi-conditional data in Section \ref{sec:exp_e2e} as the source-domain E2E model, and define LibriSpeech data , , as the target domain.
We train source-domain HAT and MHAT on 960 hours of transcribed Librispeech data \cite{panayotov2015librispeech} comprising read English speech based on LibriVox's audio books. 
We adapt MHAT using the \emph{transcripts} of Wall Street Journal (WSJ) training speech (LDC93S6B and LDC94S13B) as the adaptation text. WSJ training set consists of approximately 80 hours of read speech with text drawn from WSJ news. We evaluate the models on eval92 and eval93 consisting of 213 and 333 test utterances, respectively, and use dev93 with 503 utterances as the dev set. We train an external LM with the same WSJ transcripts used for text-only adaptation.

Then we adapt the Librispeech-trained MHAT to \emph{transcripts} of TED-LIUM training speech \cite{rousseau2012ted, hernandez2018ted}. TED-LIUM training set consists of approximately 450 hours of speech extracted from the freely available video talks from TED conferences. The dev and evaluation sets consist of 507 and 1155 TED-LIUM utterances, respectively. We train an external LM with the same TED-LIUM transcripts.

Finally, we adapt the Librispeech-trained MHAT to the \emph{transcripts} of Common Voice training speech \cite{ardila2019common}. Common Voice is a crowd-sourced dataset collected from volunteers who record sample sentences with a microphone. We use the version 5.1 (June 22 2020) snapshot with approximately 1500 hours of speech. The dev and evaluation sets both consist of 16029 Common Voice utterances. We train an external LM with the same Common Voice transcripts.

\begin{table*}[t]
\centering
\setlength{\tabcolsep}{5.5pt}
\begin{tabular}[c]{c|c||c|c|c||c|c||c|c}
	\hline
	\hline
	\multirow{2}{*}{\begin{tabular}{@{}c@{}} Method \end{tabular}} & \multirow{2}{*}{\begin{tabular}{@{}c@{}} Params \end{tabular}} & \multicolumn{3}{c||}{WSJ} & \multicolumn{2}{c||}{TED-LIUM} & \multicolumn{2}{c}{Common Voice} \\
	\hhline{~~-------}
	& & dev93 & eval92 & eval93 &  \hspace{4pt} dev \hspace{4pt} & \hspace{3pt} test \hspace{3pt} & \hspace{4pt} dev \hspace{4pt} & \hspace{3pt} test \hspace{3pt} \\
	\hline
    HAT & \multirow{2}{*}{\begin{tabular}{@{}c@{}} 128M \end{tabular}} & 12.8 & 9.5 & 12.5 & 11.4 & 10.8 & 24.2 & 27.6 \\
	\hhline{-~-------}
%	MHAT & & 11.7 & 9.0 & 11.4 \\
%	\hhline{-~---}
	MHAT & & 12.2 & 9.5 & 11.8 & 11.9 & 11.1 & 23.3 & 26.6 \\
	\hline
	\hline
	HAT + LM & \multirow{2}{*}{\begin{tabular}{@{}c@{}} 137M \end{tabular}} & 12.0 & 8.8 & 11.4 & 11.0 & 10.2 & 22.1 & 25.2 \\
	\hhline{-~-------}
	MHAT + LM & & 11.0 & 8.7 & 10.9 & 11.2 & 10.6 & 21.7 & 24.6 \\
	\hline
	\hline
	MHAT + ILMA & 128M & 10.8 & 7.7 & 9.8 & 10.8 & 10.0 & 21.0 & 24.3 \\
	\hline
	MHAT + ILMA + LM & 137M & \textbf{9.8} & \textbf{7.2} & \textbf{9.3} & \textbf{10.4} & \textbf{9.6} & \textbf{20.0} & \textbf{23.1} \\
	\hline
	\hline
	\end{tabular}
	\caption{WERs (\%) of HAT, MHAT, and MHAT after ILMA. LM fusion is performed with HAT, MHAT and internal-LM-adapted MHAT. 960h of transcribed Librispeech data is used for training. \textbf{Transcripts} of WSJ (80h), TED-LIUM (450h) and Common Voice (1500h) training data are used for respective text-only ILMA and LSTM-LM training. Internal LM scores are subtracted for HAT + LM and MHAT + LM.}
\label{table:public_ilma}
\vspace{-5 pt}
\end{table*}

\subsubsection{Modeling}
We extract 80-dim log Mel filterbank features from speech signal with a 25 ms window and a stride of 10 ms as the input. The MHAT encoder is a full-context conformer \cite{gulati2020conformer} with 17 layers and each layer has 512-dim and 8-headed self-attention.
% and 512-dimensional relative position embedding.  
The convolution module has a kernel size of 32. The label and blank decoders of MHAT are both embedding decoders \cite{botros2021tied} with embedding dimensions of 640 and 320. For each label $y_u$, we condition the decoders only on the last 2 labels $y_{u-2}$ and $y_{u-1}$ and create a $|\mathcal{V}|$-entry lookup table for each of them. The embedding vectors of $y_{u-2}$ and $y_{u-1}$ are concatenated and then projected down to the embedding dimension. The two lookup tables are independent for the label decoder and are tied for the blank decoder. The label decoder output is projected to 4095-dim output layer predicting word pieces \cite{schuster2012japanese}. We use Tanh as the non-linear function $\phi(\cdot)$ for all experiments. The label with projection $W_4$ and the blank decoder have 8.7M and 1.3M parameters, respectively, and the entire MHAT has 128M parameters. 

As a comparison, we also train a standard HAT with exactly the same input features and encoder as those of MHAT. To keep the model size consistent, we increase the embedding dimension of the HAT embedding decoder to 960. The joint network has 640 hidden units. The decoder output is projected to 4096-dim output layer predicting word pieces and a blank. HAT has in total 128M parameters with 12.2M parameters in its decoder and joint network.

The external LMs described in Section \ref{sec:public_data} are all 1-layer long short-term memories (LSTMs) \cite{sak2014long, meng2017deep, erdogan2016multi} with 640-dim hidden units and 4095 output units. The LSTM-LM has 8.7M parameters which is of the same size as the internal LM (i.e., label decoder plus $W_4$) in MHAT. We also tried fusing larger LMs, but they turned out to perform worse because transcripts from WSJ, TED-LIUM and Common Voice are rather small data and large LMs easily overfit to the training text. 

\subsubsection{MHAT Training}
In Table \ref{table:librispeech}, we compare the ASR performance of HAT and MHAT trained with the same source-domain Librispeech data. We also compare MHATs trained with different internal LM loss weights $\alpha$. MHAT trained with a HAT loss only or with $\alpha=0.1$ performs almost the same as HAT on all four Librispeech dev and test sets. Trained with $\alpha=0.1$, the MHAT internal LM has a lower token-level perplexity of 53.7 on test-clean than HAT or MHAT. This shows that MHAT with a separate AM, a standalone internal LM and a separate blank decoder is able to perform equally well as HAT. Note that the additional blank decoder only takes only 1\% of the total MHAT parameters. However, WER of MHAT begins to degrade as $\alpha$ increases to 0.5 and 1.0. To avoid this, we fix $\alpha$ at 0.1 for all other experiments. 

\begin{table}
\centering
\setlength{\tabcolsep}{3.5pt}
\begin{tabular}[c]{c|c|c||c|c|c|c}
	\hline
	\hline
	\multirow{2}{*}{\begin{tabular}{@{}c@{}} Model \end{tabular}} &  \multirow{2}{*}{\begin{tabular}{@{}c@{}} \hspace{5pt} $\alpha$ \hspace{5pt} \end{tabular}} & \multirow{2}{*}{\begin{tabular}{@{}c@{}} PPL \end{tabular}} & \multicolumn{2}{c|}{dev} & \multicolumn{2}{c}{test} \\
	\hhline{~~~----}
	& & & clean & other & clean & other\\
	\hline
    HAT & - & 142.6 & 2.1 & 4.7 & 2.2 & 4.8 \\
	\hline
	\multirow{4}{*}{\begin{tabular}{@{}c@{}} MHAT \end{tabular}} & 0.0 & 164.2 & 2.1 & 4.7 & 2.2 & 4.9 \\
	\hhline{~------}
	& 0.1 & 53.7 & 2.1 & 4.6 & 2.2 & 4.9\\
	\hhline{~------}
	& 0.5 & 53.4 & 2.1 & 4.7 & 2.3 & 5.0\\
	\hhline{~------}
	& 1.0 & 54.1 & 2.2 & 5.1 & 2.4 & 5.1\\
	\hline
	\hline
	\end{tabular}
	\caption{WERs (\%) of HAT and MHAT training with different internal LM loss weights $\alpha$. 960h transcribed Librispeech data is used for training. Perplexity (PPL) of internal LM is measured on test-clean.}
\label{table:librispeech}
\vspace{-10 pt}
\end{table}

\subsubsection{ILMA vs. LM Fusion}
% We compare the ASR performance of MHAT with ILMA to baseline unadapted models and to LM fusion methods. 
For cross-domain experiments, we adapt Librispeech-trained source-domain MHAT to each of the three target domains including WSJ, TED-LIUM and Common Voice, using their respective \emph{transcripts} of training data.
% and evaluate on their respective test sets. 
In Table \ref{table:public_ilma}, we first evaluate baseline HAT and MHAT on three target-domain dev and test sets. Although HAT and MHAT achieves similar WERs on Librispeech, 
% HAT and MHAT perform differently on three target-domain datasets. 
MHAT performs better than HAT on WSJ and Common Voice but worse on TED-LIUM. We further integrate each of the three external LMs trained with three target-domain transcripts, respectively, into HAT and MHAT. Note that we subtract the internal LM scores during the LM fusion \cite{variani2020hybrid,meng2021ilme}. Both HAT and MHAT get improved ASR performance. MHAT with LM fusion achieves 1.8\%--14.1\% and 3.3\%--7.7\% relative WER reductions from the baseline HAT and MHAT over all three target domains.
% 3.3\%-6.0\%, 6.2\%-6.7\% and 6.5\%-7.7\% relative WER reductions from baseline MHAT on WSJ, TED-LIUM and Common Voice, respectively. 

We adapt the internal LM of MHAT to each of the three target-domain transcripts, respectively. MHAT with ILMA achieves 5.3\%--21.6\% and 7.6\%--14.4\% relative WER reductions from the baseline HAT and MHAT over all three target domains. Correspondingly, internal LM perplexities of MHAT are reduced from 362.9, 284.0 and 487.8 to 57.3, 58.0 and 51.4, respectively, after ILMA on WSJ eval92, TED-LIUM and Common Voice test sets, respectively. This explains the significantly lower WERs achieved by ILMA-ed MHAT: after ILMA, MHAT AM and blank decoder are able to work with a much better internal LM with greatly reduced perplexity on the target domain.
% 7.7\%-14.4\%, 10.0\%-11.5\%, and 7.6\%-10.6\% relative WER reductions from baseline MHAT, respectively, on the three datasets, respectively. 
% Note that MHAT with ILMA consistently outperforms HAT with LM fusion by 1.8\%--14.0\%, relatively, over all three target domains, despite having 7\% fewer model parameters. 
Note that MHAT with ILMA consistently outperforms MHAT with LM fusion by 1.8\%--11.5\%, relatively, over all three target domains, despite having 7\% fewer model parameters. 

More importantly, additional 3.7\%--9.3\% relative WER reductions from ILMA-ed MHAT are achieved by further integrating the same external LM into MHAT, leading to in total 8.8\%--25.6\% and 12.2\%--20.0\% relative WER reductions from the baseline HAT and MHAT, respectively, over all three target domains. This shows that the two text-only adaptation methods, ILMA and external LM fusion, are complementary. Adapting the internal LM of MHAT to target-domain text facilitates a more effective fusion with an external LM of the same domain, achieving much better ASR performance than HAT/MHAT with ILME-based fusion.

\begin{table*}[h]
\centering
\setlength{\tabcolsep}{5.5pt}
\begin{tabular}[c]{c|c|c|c|c|c|c|c}
	\hline
	\hline
	\multirow{2}{*}{\begin{tabular}{@{}c@{}} Method \end{tabular}} & \multirow{2}{*}{\begin{tabular}{@{}c@{}} Params \end{tabular}} & \multirow{2}{*}{\begin{tabular}{@{}c@{}} \hspace{0pt} Voice \hspace{0pt}\\ \hspace{0pt} Search \hspace{0pt} \end{tabular}} & \multicolumn{5}{c}{Rare Words} \\
	\hhline{~~~-----}
	& & & \hspace{4pt} Maps \hspace{4pt} & \hspace{6pt} Play \hspace{6pt} & \hspace{6pt} Web \hspace{6pt} & \hspace{-2pt} YouTube \hspace{-2pt} & \hspace{5pt} SxS \hspace{5pt} \\
	\hline
    HAT & \multirow{2}{*}{\begin{tabular}{@{}c@{}} 155M \end{tabular}} & 5.9 & 13.0 & 37.1 & 21.2 & 24.1 & 24.9 \\
	\hhline{-~------}
	MHAT & & 6.0 & 12.8 & 36.7 & 20.1 & 23.3 & 23.4 \\
	\hline
	\hline
	HAT + LM & \multirow{2}{*}{\begin{tabular}{@{}c@{}} 198M \end{tabular}} & 5.9 & 11.5 & 34.9 & 17.4 & 22.1 & 23.0 \\
	\hhline{-~------}
	MHAT + LM & & \textbf{5.8} & 11.7 & 34.9 & 17.9 & 21.9 & 22.1 \\
	\hline
	\hline
	MHAT + ILMA & 155M & 6.0 & 11.9 & 34.8 & 18.6 & 21.6 & 22.6 \\
	\hline
	MHAT + ILMA + LM & 198M & 5.9 & \textbf{11.0} & \textbf{33.6} & \textbf{16.7} & \textbf{20.5} & \textbf{20.9} \\
	\hline
	\hline
	\end{tabular}
	\caption{WERs (\%) of HAT, MHAT, and ILMA-ed MHAT.
	LM fusion is further conducted with all 3 models. 
	400K hours of multi-domain transcribed speech is used for training. 100B multi-domain sentences are used for respective text-only ILMA and LSTM-LM training. Internal LM scores are subtracted for HAT + LM and MHAT + LM.}
\label{table:prod_ilma}
\vspace{-5 pt}
\end{table*}

\subsection{Multi-Domain Adaptation}
\label{sec:md_adapt}
\subsubsection{Dataset}
\label{sec:production_data}
In Section \ref{sec:md_adapt}, all E2E models are trained with 400K hours of multi-domain English data comprising $\sim$300M audio-transcript pairs as in \cite{sainath2022improving, narayanan2019recognizing}. The training data covers multiple domains including voice search, farfield use cases, segmented telephony speech, and YouTube video segments. The amount of audio in each domain is detailed in \cite{narayanan2019recognizing}. YouTube transcripts are generated in a semi-supervised fashion \cite{liao2013large} while data from other domains is anonymized and hand-transcribed adhering to Google AI principles \cite{googleai}. In addition, multi-condition training \cite{kim2017generation}, random 8kHz down-sampling \cite{li2012improving} and SpecAug \cite{park2019specaugment} are applied to augment and diversify the data.

We use large-scale multi-domain text-only data for adaptation and external LM training. The text-only data consists of 100B anonymized sentences across the domains of Maps, Google Play, Web, and YouTube.
%, for better rare-word recognition.
% selected from 213B anonymized search traffic 
% We hope to improve the rare word recognition in these domains through text-only adaptation. 

We evaluate our models on the Voice Search test set containing around 12K anonymized and hand-transcribed voice search utterances with an average duration of 5.5 seconds. To measure ASR performance on long-tail words, we also evaluate our model on both TTS-generated speech and real speech containing rare words. One TTS test set is used for each of the four domains: Maps, Google Play, Web and YouTube. All TTS sets include rare proper nouns that appear fewer than 5 times in the training set. In addition, we use a real side-by-side (SxS) test set \cite{sainath2019two} including 1K utterances on which an E2E model produces more recognition errors than a state-of-the-art conventional model. With text-only adaptation, our goal is to improve the ASR accuracy on the five rare-word test sets without degrading the WER on Voice Search.
\vspace{-1pt}

\subsubsection{Modeling}
From the speech signal, we extract 128-dim log Mel filterbank features with a 32 ms window and a 10 ms stride. Four adjacent features are stacked to form a 512-dim feature, which is then down-sampled to a frame rate of 30 ms. We append each speech feature with a 16-dim domain identifier before passing it to MHAT. MHAT has 2-pass cascaded encoders \cite{narayanan2021cascaded}: the 1st pass causal encoder processes the input speech features and the 2nd pass non-causal encoder operates on the output of the causal encoder. Both label and blank decoders of a cascaded MHAT have to decode either using the output of the causal or non-causal encoder. The causal encoder has 7 conformer layers with only left-context attention to prevent the model from accessing future inputs. The non-causal encoder has 10 conformer layers with additional right-context attention processing 900 ms of speech into the future. A 512-dim self-attention with 8 heads and a convolution kernel of size 15 are used in each conformer layer.
To learn from much larger amount of multi-domain text, we increase the capacity of the label decoder to a 2-layer LSTM with 1600 hidden units in each layer. The blank decoder remains to be an embedding decoder with an embedding dimension of 320 and with a look-up table shared between $y_{u-2}$ and $y_{u-1}$. The label decoder output is projected to 4095-dim output layer corresponding to the same word pieces. The label and blank decoder have 43M and 1.3M parameters, respectively, and the entire cascaded MHAT has 155M parameters. During ILMA, the KL regularization weight $\rho$ is set to 0.5 since we want to keep source-domain Voice Search performance unchanged.
% All models are trained with FastEmit in boths passes.

As a comparison, we also train a cascaded HAT with exactly the same input features and causal encoder as those of MHAT. HAT has an embedding label decoder with an embedding dimension of 640. The cascaded HAT has in total 155M parameters.
To match the size of MHAT internal LM, the external LM described in Section \ref{sec:production_data} is a 2-layer LSTM with 1600-dim hidden units in each layer and with 4095 output units. The LSTM-LM has in total 43M parameters.

\subsubsection{Results}
%We compare the ASR performance of MHAT ILMA with baseline unadapted models and with LM fusion methods. 
Given limited space, we only show WERs with the 2nd-pass non-causal encoder in this paper. The trend of the 1st-pass WERs is very similar. When trained with 400K-hour multi-domain data, MHAT performs slightly better than HAT on rare-word test sets, but slightly worse on Voice Search as shown in Table \ref{table:prod_ilma}.
We then integrate an external LM trained with 100B multi-domain sentences into HAT and MHAT, respectively. Note that we subtract the internal LM scores during LM fusion. Both HAT and MHAT get reduced WERs. MHAT with LM fusion achieves 1.8\%--15.5\% and 5.8\%--10.9\% relative WER reductions from the baseline HAT and MHAT over all rare-word test sets.

We adapt the internal LM of MHAT to the same 100B multi-domain sentences. MHAT with ILMA achieves 6.3\%--12.4\% and 4.4\%--7.1\% relative WER reductions from the baseline HAT and MHAT over all TTS test sets. The result again shows the effectiveness of ILMA on adapting MHAT to text-only data. 
%MHAT with ILMA performs 9.7\% relatively better than baseline HAT in terms of lower WER, but 1.4\% worse than baseline MHAT. 
Compared to HAT/MHAT with LM fusion, MHAT with ILMA performs better on YouTube and Google Play, but worse on Maps and Web. This is still reasonable because ILMA-ed MHAT has 21.7\% fewer run-time parameters than LM fusion. Further, with additional external LM fusion, the ILMA-ed MHAT manages to achieve 9.6\%--21.5\% relative WER reductions from the baseline HAT and 6.1\%--16.7\% relative WER reductions from the baseline MHAT. ILMA-ed MHAT with LM fusion consistently outperforms HAT with LM fusion and MHAT with LM fusion by relatively 3.9\%--9.2\% and 3.8\%--7.1\%, respectively. This again implies that, for LM fusion, adapting the internal LM of MHAT to target domains is more effective than subtracting the source-domain internal LM from HAT. 
% 7.7\%-14.4\%, 10.0\%-11.5\%, and 7.6\%-10.6\% relative WER reductions from baseline MHAT, respectively, on the three datasets, respectively. 
% ILMA can better benefit the external LM integration than the internal LM subtraction.  
As expected, ILMA with KL regularization does not change the Voice Search performance too much.

\section{Conclusions}
 In this work, we propose a novel modular HAT that has separate label and blank decoders performing label and blank predictions, respectively, on top of a shared acoustic encoder. We project the encoder and label decoder outputs directly to AM and internal LM scores with label dimensions and add them together to estimate the label posteriors. We are able to train MHAT with an additional internal LM loss such that the label decoder becomes an standalone LM that is adaptable to text-only data by minimizing a simple cross-entropy loss. MHAT with ILMA achieves up to 21.6\% and 12.4\% relative WER reductions from baseline HAT on public datasets and production datasets with long-tail words, respectively. Additional gains are achieved by further integrating an external LM into the ILMA-ed MHAT, leading up to 25.6\% and 21.5\% relative WER reductions from the baseline HAT on public and internal production sets, respectively.

\newpage

\bibliographystyle{IEEEbib}
\bibliography{refs}

\end{document}